\documentclass[runningheads]{llncs}
\usepackage[T1]{fontenc}
\usepackage{graphicx}
\usepackage{multirow}
\usepackage{algorithm}
\usepackage{algpseudocode}
\usepackage{amsmath}
\usepackage[misc,geometry]{ifsym}
\usepackage{xcolor}

\begin{document}
\title{The \textit{M-factor}: A Novel Metric for Evaluating Neural Architecture Search in Resource-Constrained Environments}
%
%

\author{Srikanth Thudumu\inst{1}\Letter\orcidID{0000-0002-7848-9008} \and
Hy Nguyen \inst{1}\orcidID{0009-0008-7092-3106}  \and
Hung Du \inst{1}\orcidID{0000-0003-1415-5786} \and 
Nhat Duong \inst{1}\orcidID{0009-0005-1349-8666} \and
Zafaryab Rasool \inst{1}\orcidID{0000-0002-3603-3125} \and
Rena Logothetis \inst{1}\orcidID{0000-0003-0143-3812} \and
Scott Barnett \inst{1}\orcidID{0000-0002-3187-4937} \and
Rajesh Vasa\inst{1}\orcidID{0000-0003-4805-1467} \and
Kon Mouzakis\inst{1}\orcidID{0000-0003-4447-5166}
}
\authorrunning{S. Thudumu et al.}
%
\institute{Applied Artificial Intelligence Institute ($A^2I^2$), Deakin University, Geelong VIC 3216, Australia 
\email{\{Srikanth.Thudumu, Hy.Nguyen, Hung.Du, Nhat.Duong, Zafaryab.Rasool, Rena.Logothetis, Scott Barnett, Rajesh.Vasa, Kon.Mouzakis\}@deakin.edu.au}
\\
}

\maketitle              
\begin{abstract}
Neural Architecture Search (NAS) aims to automate the design of deep neural networks. However, existing NAS techniques often focus primarily on maximizing accuracy, neglecting model efficiency. This limitation hinders their applicability in resource-constrained environments such as mobile devices and edge computing systems. Additionally, current evaluation metrics typically prioritize performance over efficiency, lacking a balanced approach to assess architectures suitable for deployment in constrained scenarios. To address these limitations, this paper introduces the \textit{M-factor}, a novel metric that combines model accuracy and size. We compare four diverse NAS techniques: Policy-Based Reinforcement Learning, Regularized Evolution, Tree-structured Parzen Estimator (TPE), and Multi-trial Random search. This selection represents different approaches in NAS, allowing for a comprehensive assessment of the M-Factor across various paradigms. The study examines ResNet configurations on the CIFAR-10 dataset, with a search space of 19,683 configurations. Experiments show Policy-Based Reinforcement Learning and Regularized Evolution achieved \textit{M-factor} values of 0.84 and 0.82 respectively, while Multi-trial Random search attained 0.75 and TPE reached 0.67. Policy-based reinforcement Learning exhibited performance changes after 39 trials, and Regularized Evolution showed optimization within 20 trials. The research analyzes optimization dynamics and trade-offs between accuracy and model size for each strategy. Results indicate that in some cases, random search performed comparably to more complex algorithms when evaluated using the \textit{M-factor}. These findings demonstrate how the \textit{M-factor} addresses the limitations of existing metrics by guiding NAS towards balanced architectures, providing insights into strategy selection for scenarios requiring both model performance and efficiency.




\end{abstract}
\section{Introduction}
Deep neural networks have achieved remarkable success across various domains; however, their increasing complexity leads to computationally expensive and memory-intensive models \cite{lyu2024survey}. This creates challenges for deployment in resource-constrained environments such as mobile devices, edge computing systems, and IoT applications \cite{howard2017mobilenets}. Neural Architecture Search (NAS) addresses these challenges using an automatic technique to optimise neural network architectures with an efficient number of hyperparameters resulting in a high accuracy. However, existing NAS techniques primarily focus on maximizing accuracy, often neglecting the crucial aspect of model efficiency in terms of the number of parameters \cite{elsken2019neural,zoph2017neural}. 


To support model development for resource-constrained environments, we introduce \textit{M-Factor}, a novel metric that measures the trade-off between performance and efficiency in NAS. Our study focuses on four distinct NAS techniques: Policy-based Reinforcement Learning \cite{zoph2017neural,zoph2018learning}, Regularized Evolution \cite{real2019regularized}, Tree-structured Parzen Estimator (TPE) \cite{bergstra2011algorithms,bergstra2013making}, and Multi-trial Random search \cite{li2020random}. We apply these methods to optimize ResNet architectures \cite{he2016deep} for the CIFAR-10 dataset \cite{krizhevsky2009learning}, using the \textit{M-Factor} as our primary evaluation metric.

Our work contributes to the ongoing effort to develop efficient and effective methods for automated neural architecture design, particularly for resource-constrained environments. By utilizing our proposed metric for comparing these 
diverse NAS approaches, we provide insights in optimizing the trade-off between performance and efficiency. It is important to note that this study did not implement gradient-based search techniques such as DARTS \cite{liu2018darts}, which use continuous and differentiable architecture representations, due to framework constraints in accommodating our custom \textit{M-Factor} metric. Weight sharing strategies such as ENAS and one-shot models \cite{pham2018efficient,bender2018understanding}, which train a single over-parameterized model containing all possible sub-network architectures, were also excluded. Our experimental setup precluded the evaluation of sub-networks within a larger model. The selection of techniques was influenced by available computational resources, favoring methods that could run effectively within our constraints. Our search space, focused on specific layers within ResNet blocks, also guided our choice of search strategies.The main contributions of this paper are:
\begin{itemize}
    \item Development of \textit{M-Factor}, a customized metric that enables NAS to optimize the trade-off between accuracy and model size.
    \item Comparison of four diverse NAS techniques, providing insights into their effectiveness in optimizing the \textit{M-Factor}.
    \item Analysis of the trade-offs between accuracy and model size achieved by different NAS strategies, offering practical insights for deploying models in resource-constrained environments.
\end{itemize}

The structure of the paper is as follows: Section 2 presents background on architecture search and the specific techniques we focus on. Section 3 discusses the proposed \textit{M-Factor} metric. Section 4 outlines our experimental setup, including the dataset, network architecture, and methodology. Section 5 presents our results and provides a detailed discussion of our findings. Finally, Section 6 concludes the paper and suggests directions for future work.

\section{Related work}

Due to the demand of small deep learning models operating on mobile devices or edge computing systems, research in NAS has rapidly progressed \cite{ren2021comprehensive,chitty2023neural,kang2023neural,10.1145/3665138}. The goal of NAS is to identify the optimal neural network architectures that have an efficient number of hyperparameters to attain high accuracy. Zoph and Le presented an initial popular work in this area \cite{zoph2016neural} based on reinforcement learning (RL). Since then, a number of works have focused on different approaches for optimal deep neural network design. Apart from RL-based approaches \cite{zoph2018learning,pham2018efficient,baker2016designing}, other approaches \cite{miikkulainen2024evolving,liu2017hierarchical} are based on evolutionary algorithms or are heuristic-based \cite{liu2018progressive}.
Some existing works focused on addressing computational issues and huge memory requirements of these architectures. For example, Mellor et. al.\cite{mellor2021neural} proposed an efficient NAS algorithm. Zhang et. al.\cite{zhang2020memory} proposed a memory-efficient NAS for image denoising. Lopes et. al.\cite{lopes2024manas} proposed two lightweight implementations for NAS using a multi-agent framework to reduce memory requirements and achieve better performance.


In evaluating these NAS techniques, various general metrics are employed to assess individual aspects of performance and efficiency \cite{canziani2016analysis}. Common metrics include accuracy (A), which measures the model's performance on specific tasks, and computational costs like FLOPs (Floating Point Operations) and inference time, which gauge the efficiency of the architecture. However, these metrics suffer from the trade-off between each other. 

To provide a more holistic view, composite metrics have been developed that integrate multiple factors into a single score \cite{tan2019efficientnet,wu2019fbnet,cai2019proxylessnas,elsken2019neural,howard2017mobilenets,fang2019tinier}. These metrics combine performance indicators with efficiency measures, offering a more comprehensive evaluation of architectures. Especially, in resource-constrained scenarios, the integration of different metrics support to balance the trade-off between them. For example, Accuracy-to-Parameter Ratio (A/P) combines accuracy with the number of parameters and helps assess how efficiently a model achieves high accuracy relative to its complexity.



Similarly, Accuracy-to-FLOPs (A/F) Ratio evaluates how effectively a model achieves high accuracy for a given amount of computation, making it useful for understanding the trade-off between performance and computational efficiency.


Wong \cite{wong2019netscore} proposed NetScore that assess the performance of neural network architecture for practical usage. It aggregates accuracy, model size, and computational cost to reflect both effectiveness and efficiency. It is defined as: 
\begin{equation}
    \epsilon = 20\log\left( \frac{A^{\alpha}}{P^{\beta} \times M^{\gamma}} \right)
\end{equation}
where $M$ is the number of multiply-accumulate (MAC) operations, $\alpha$, $\beta$ and $\gamma$ are coefficients that control the impact of each metric in on the overall performance. However, the value of $P$ and $M$ increase exponentially once the complexity of the architecture increases. Furthermore, $\epsilon$ is not bounded within any range, and hence, challenging to control the upper bound of the overall performance.

To simplify these, we propose \textit{M-Factor} that measure the trade-off between the accuracy and the model efficiency in terms of its size as follows:
\begin{equation}
    M_{\alpha} = \frac{(1 + \alpha) \times A \times S'}{(\alpha \times A) + S'}
\end{equation}
Details of this equation is provided in Section \ref{sec:our_metric}.

\section{Problem Preliminary}
Neural Architecture Search (NAS) automates the process of finding neural network architectures that balance accuracy with computational and memory efficiency \cite{elsken2019neural}. It defines a search space of architectural dimensions and uses an optimizer to explore this space, guided by performance metrics. The search space denoted by $\mathcal{A}$ in NAS includes choices such as the number of layers, types of layers (e.g., convolutional, recurrent, fully connected, and other types), and hyperparameters for each layer (e.g., kernel size, stride, number of filters, and other hyperparameters):
\begin{equation}
    \mathcal{A}=\left\{\text{Arc}_1, \text{Arc}_2, \ldots, \text{Arc}_n\right\}
\end{equation}
where $\text{Arc}_i$ represents a potential architecture. Furthermore, NAS consists of a search strategy that determines how the algorithm searches for the optimal architecture as:
\begin{equation} \label{eq:search_A}
    \text{Arc}^*=\operatorname{SearchStrategy}(\mathcal{A}, \mathcal{F}, \mathcal{C})
\end{equation}
where $\mathcal{F}$ is a performance evaluation function and $\mathcal{C}$ is a cost function. Notably, the performance of each architecture is estimated by utilizing the validation dataset from the original one. In addition, $\mathcal{C}$ in the resource-constrained scenarios depends on the goal of optimization, such as reducing number of parameters, floating-point operations per second (FLOPs) or latency. The goal of NAS is to maximize the model performance while adhering to specified constraints, and Equation \ref{eq:search_A} can be extended as follows:
\begin{equation}
    \text{Arc}^*=\arg \max_{\text{Arc}_i \in \mathcal{A}} \mathcal{F}\left(\text{Arc}_i ; D_{\text {val}}\right) \text { subject to } \mathcal{C}\left(\text{Arc}_i\right) \leq \theta
\end{equation}
where $D_{\text {val}}$ is the validation dataset, and $\theta$ is a threshold for the cost.

NAS can add significant computational overhead during initial exploration \cite{zoph2017neural}. This requires a careful search space design that ensures discoveries meet latency, energy consumption, and form factor requirements. There are also challenges in reproducing optimizations across different software and hardware stacks. Despite these challenges, NAS holds promise for finding performant model architectures tailored to specific efficiency goals. NAS algorithms explore the vast space of possible neural network architectures, evaluating candidate models based on their performance on a validation set or a surrogate objective function \cite{elsken2019neural}. This search process can be formulated as an optimization problem, where the goal is to find the architecture that maximizes the desired performance metric, such as accuracy or efficiency. In our study, we focus on four specific NAS techniques:

\begin{enumerate}
    \item Reinforcement Learning (RL) methods frame architecture search as a Markov Decision Process, where an agent (controller) sequentially constructs architectures by sampling from a set of operations \cite{zoph2017neural,pham2018efficient,zoph2018learning,baker2022designing,liu2018progressive}. The controller is trained using policy gradients to maximize the expected accuracy of sampled architectures on a validation set. In our implementation, we use Policy-based RL, which optimizes the policy to maximize our custom metric \textit{M-Factor} that balances accuracy and model size.

    \item Evolutionary algorithms such as Regularized Evolution treat neural architectures as individuals in a population that evolve over generations through genetic operations like mutation and crossover \cite{real2019regularized,real2020automl}. The fitness of each architecture is evaluated on a validation set, and the fittest architectures are selected to produce the next generation. We implement Regularized Evolution in our study, using our custom metric \textit{M-Factor} as the fitness function.
    \item Tree-structured Parzen Estimator (TPE) is a sample-based approach that uses Bayesian optimization to guide the search process \cite{bergstra2011algorithms,bergstra2013making}. It builds a probabilistic model of the relationship between architectural choices and the resulting performance. We include TPE in our study to explore its effectiveness with our custom metric and search space.
    \item Multi-trial Random Search serves as a simple yet often effective baseline in NAS studies \cite{bergstra2012random,li2020random}. It randomly samples architectures from the search space for evaluation. We include this method to provide a benchmark for comparing more sophisticated strategies.
\end{enumerate}

\section{Proposed Metric: \textit{M-factor}} \label{sec:our_metric}

\subsection{Motivation}

In the context of Neural Architecture Search (NAS), traditional approaches focus primarily on maximizing model accuracy. However, in many real-world applications, particularly those involving resource-constrained environments such as mobile devices or edge computing systems, model efficiency is equally crucial. The challenge lies in finding an optimal balance between model performance and resource utilization. To address this, we propose the \textit{M-Factor}, a novel metric designed to guide NAS algorithms towards architectures that achieve high accuracy while maintaining efficiency in terms of model size. 

\subsection{Formulation}

Inspired by F1-score that utilize the harmonic mean \cite{ferger1931nature} to evaluate the balance between recall and precision, \textit{M-Factor} is designed to assess the trade-off between a model's accuracy and its efficiency in terms of size. The metric is defined as follows:
\begin{equation}
    M = \frac{2 \times A \times S'}{A + S'}
\end{equation}
where $A$ is the model accuracy (on validation set), and $S'$ is a normalized inverse measure of the model size:
\begin{equation} \label{eq:s}
    S' = \frac{P_{min}}{P}
\end{equation}
where $P$ is the number of parameters in the current model, and $P_{min}$ is the number of parameters in the smallest model in the search space. There are two reasons for the design of $S'$ as in Equation \ref{eq:s}:
\begin{enumerate}
    \item $S'$ is the inverse of $P$. This ensures that the harmonic mean when optimizing $S'$ is equivalent to minimizing $P$.
    \item We have $P_{min}$ divided by $P$ to ensure the range of $S'$ lies between 0 and 1. Since the range of $A$ is also 0 to 1, two elements such as the accuracy and the model size should be in the same range. This aims to eliminate the dominance of one element on the other element in the harmonic mean.
\end{enumerate}
Notably, the harmonic mean is sensitive to low values, meaning that if either model size or accuracy is low, \textit{M-factor} will be significantly lower than their arithmetic mean. Consequently, \textit{M-factor} only achieves a high value when both metrics are high. This property makes it effective for ensuring that models are optimal in terms of both size and accuracy.


\subsection{Weighted Variant}

Aside from balancing between the model accuracy and its size, many resource-constrained scenarios require to prioritize one of these factors. To attain this, we introduce a weighted variant of \textit{M-Factor} as follows:
\begin{equation}
    M_{\alpha} = \frac{(1 + \alpha) \times A \times S'}{(\alpha \times A) + S'}
\end{equation}
where $\alpha$ is the weight factor that controls the relative importance of accuracy versus model size. There are three potential cases for the value of $\alpha$ as:
\begin{enumerate}
    \item When $\alpha = 1$, we get the original, balanced \textit{M-factor}.
    \item As $\alpha$ is greater than 1 ($\alpha > 1$), more emphasis is placed on minimizing model size.
    \item As $\alpha$ varies from 0 to 1  ($0 < \alpha < 1$), the accuracy is prioritized over the model size.
\end{enumerate}
This adaptability makes the \textit{M-factor} particularly valuable for scenarios with varying resource constraints or different priorities between performance and efficiency.

\section{Experiments}

\subsection{Dataset}
Our experiments were conducted using the CIFAR-10 dataset \cite{krizhevsky2009learning}. This dataset comprises 60,000 32x32 color images distributed across 10 classes, with 6,000 images per class. The dataset is pre-divided into 50,000 training images and 10,000 test images.

\subsection{Search Space}
We based our experiments on the ResNet architecture \cite{he2016deep}, with a specific focus on modifying the convolutional layers within ResNet blocks. Each ResNet block in our experiments contained a \texttt{conv1} layer, and we define three different \texttt{LayerChoice} options for that \texttt{conv1} layer:

\begin{enumerate}
    \item 2D convolution with kernel size 3x3, stride=stride, padding=1
    \item 2D convolution with kernel size 5x5, stride=stride, padding=2
    \item 2D convolution with kernel size 7x7, stride=stride, padding=3
\end{enumerate}

All convolution layers were configured with \texttt{bias=False}. We designed our ResNet model to have three main layers, each main layer has three ResNet blocks, and each ResNet block has one \texttt{conv1} layer. Therefore, we have nine \texttt{conv1} layers. Because each \texttt{conv1} layer has three options, hence we have $3^9=19,683$ total configurations to search. We also make sure that the ResNet blocks are not removed during the searching process. This helps to keep the value of $P_{min}$ in Equation 2 unchanged and prevent it from being reduced to 0.

Our experiment was designed to test 9 different combinations of these \texttt{LayerChoice} options across 3 ResNet blocks. This setup allowed us to systematically explore the impact of varying kernel sizes on the network's performance. The number of input and output planes for each convolution layer was kept consistent within a block but could vary between blocks, as indicated by the \texttt{in\_planes} and \texttt{planes} parameters in the configuration.

\subsection{Implementation Details}
The experiment was implemented using PyTorch \cite{paszke2019pytorch}, as evidenced by the use of \texttt{nn.LayerChoice} and \texttt{nn.Conv2d} modules in the configuration code. By systematically varying these convolutional layer configurations, we aimed to investigate their effects on model performance, including aspects such as accuracy, training speed, and generalization capability on the CIFAR-10 dataset.

\subsection{Experimental Methodology}

As mentioned above, we focused our experimental methodology on comparing Neural Architecture Search (NAS) strategies for ResNet models on the CIFAR-10 dataset. Our search space comprised three layers, each layer has three ResNet blocks, each with three LayerChoice options for the conv1 layer (kernel sizes 3$\times$3, 5$\times$5, and 7$\times$7), resulting in $3^9=19,683$ possible architectures. To evaluate model performance, we developed a custom metric \textit{M-Factor} that balanced accuracy and model size, allowing for flexible prioritization through a weight factor $\alpha$. We implemented and compared four NAS strategies: Policy-based Reinforcement Learning (RL), Regularized Evolution, Tree-structured Parzen Estimator (TPE), and Multi-trial Random as a baseline.

We conducted 50 trials for each strategy to ensure fair comparison. Throughout the experiments, we tracked the best result (highest \textit{M-Factor} value) achieved, plotted optimization curves, and generated accuracy and model size graphs over trials. We also analyzed the performance of the top 20\% models for selected strategies. Our comparative analysis included evaluating the best results achieved by each strategy, analyzing optimization speed and consistency, and assessing the trade-offs between accuracy and model size. We also considered practical limitations, such as our inability to implement one-shot strategies like DARTS and ENAS due to framework constraints. This comprehensive approach allowed us to thoroughly compare NAS strategies, providing insights into their effectiveness and behavior in the context of ResNet architectures on the CIFAR-10 dataset.

\section{Results and Discussion}

\subsection{Performance of Search Strategies}

We evaluated four Neural Architecture Search (NAS) strategies on the CIFAR-10 dataset using our custom metric \textit{M-factor}. The best results achieved by each strategy are summarized in Table \ref{tab:strategy_comparison}.

\begin{table}[h]
\centering
\begingroup
\renewcommand{\arraystretch}{2}
\begin{tabular}{|l|c|c|c|}
\hline
Strategy & Best \textit{M-Factor} & Accuracy & Number of parameters \\
\hline
Policy-based RL & 0.84 & 0.7637 & 284,762\\
Regularized Evolution & 0.82 & 0.755 & 301,146 \\
Multi-trial Random & 0.75 & 0.6799 & 401,498\\
TPE & 0.67 & 0.7338 & 350,298\\
\hline
\end{tabular}
\endgroup
\caption{Comparison of NAS strategies based on best \textit{M-Factor} value achieved}
\label{tab:strategy_comparison}
\end{table}

Policy-based Reinforcement Learning (RL) achieved the highest \textit{M-Factor} value of 0.84, demonstrating its effectiveness in navigating the search space to find architectures that balance accuracy and model size. This result suggests that the RL approach is capable of learning and adapting its search strategy over time to optimize for our custom metric. Regularized Evolution closely followed with an \textit{M-Factor} value of 0.82, indicating that evolutionary algorithms can also be highly effective for this NAS task. The small difference between Policy-based RL and Regularized Evolution (0.02) suggests that both methods are competitive in this search space. Interestingly, the Multi-trial Random strategy outperformed TPE with an \textit{M-Factor} value of 0.75. This result underscores the importance of including simple baselines in NAS experiments, as random search can sometimes be more effective than more complex algorithms, especially in smaller search spaces. The Tree-structured Parzen Estimator (TPE) achieved the lowest \textit{M-Factor} value of 0.67, which was unexpected given its sophistication. This outcome suggests that TPE may not be well-suited for our specific search space or may require further tuning to perform optimally with our custom metric.

\subsection{Optimization Dynamics}

\begin{figure}[h]
    \centering
    \includegraphics[width=1\textwidth]{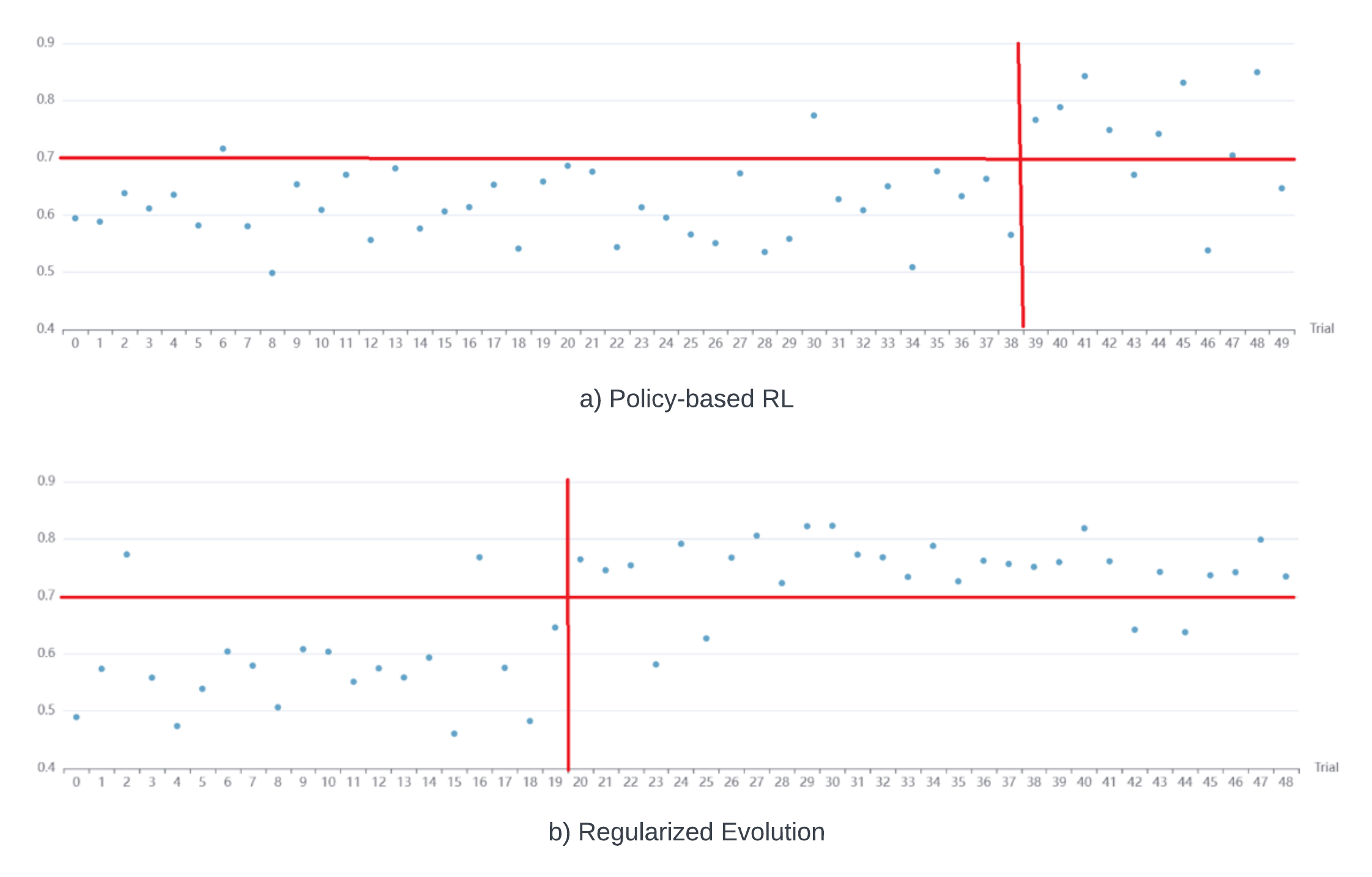}
    \caption{The M-factor values of Policy-based RL and Regularized Evolution techniques over 50 trials. Red lines indicate clear improvement trends.}
    \label{fig:RL_and_Reg}
\end{figure}

\begin{figure}[h]
    \centering
    \includegraphics[width=1\textwidth]{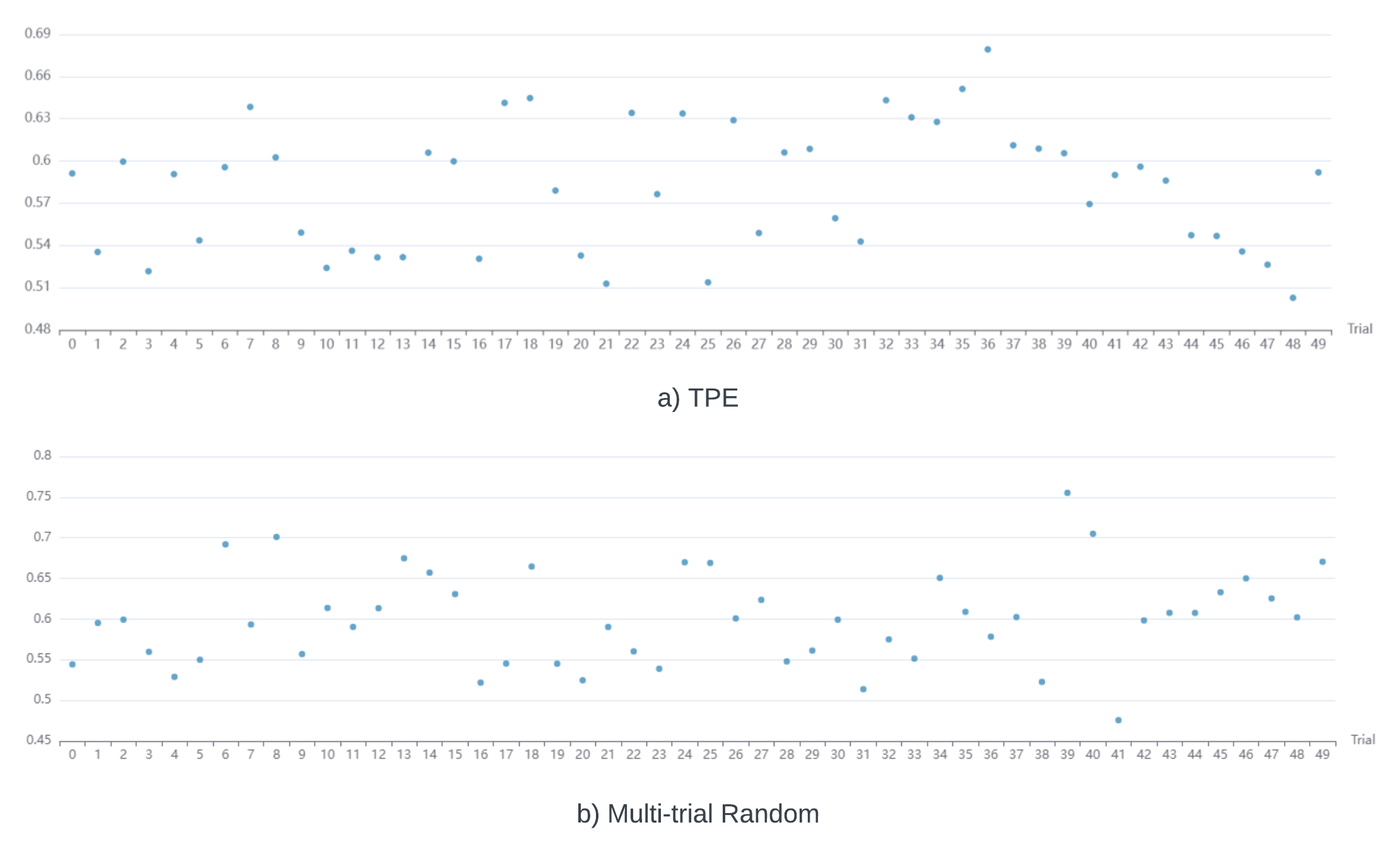}
    \caption{The \textit{M-factor} values of TPE and Multi-trial Random. There is no clear improvement trend for these two techniques.}
    \label{fig:tpe_and_random}
\end{figure}

The results of the M-factor across 50 trials revealed distinct behaviors for each strategy, providing insights into their search processes:

\begin{enumerate}
    \item \textbf{Policy-based RL:} This strategy showed a notable improvement after the 39th trial, with \textit{M-Factor} values consistently above 0.7 thereafter (Figure \ref{fig:RL_and_Reg}a). This behavior indicates that the RL agent required a significant number of trials to learn an effective policy for navigating the search space. Once learned, however, the policy consistently produced high-performing architectures. This suggests that Policy-based RL might be particularly effectivee for longer-running NAS experiments where the initial learning period can be amortized.

    \item \textbf{Regularized Evolution:} Demonstrating faster initial optimization compared to Policy-based RL, Regularized Evolution reached improved performance after just 20 trials (Figure \ref{fig:RL_and_Reg}b). This rapid improvement indicates that the evolutionary approach quickly identified and propagated beneficial architectural traits. The strategy's ability to find good solutions early makes it potentially more suitable for scenarios with limited computational resources or time constraints.

    \item \textbf{TPE:} The optimization curve for TPE exhibited inconsistent performance without a clear improvement trend (Figure \ref{fig:tpe_and_random}a). This behavior suggests that TPE struggled to build an effective probabilistic model of the search space with respect to our custom metric. The lack of consistent improvement over time indicates that TPE might not be well-suited for our specific combination of search space and evaluation metric.

    \item \textbf{Multi-trial Random:} While this strategy doesn't have a learning pattern, it provided a strong baseline (Figure \ref{fig:tpe_and_random}b). Its performance underscores the importance of comparing against simple strategies in NAS experiments. The fact that it outperformed TPE highlights that in some cases, especially with smaller search spaces, the complexity of more sophisticated algorithms may not translate to better performance.
\end{enumerate}

\subsection{Trade-offs Analysis}

\begin{figure}[h]
    \centering
    \includegraphics[width=1\textwidth]{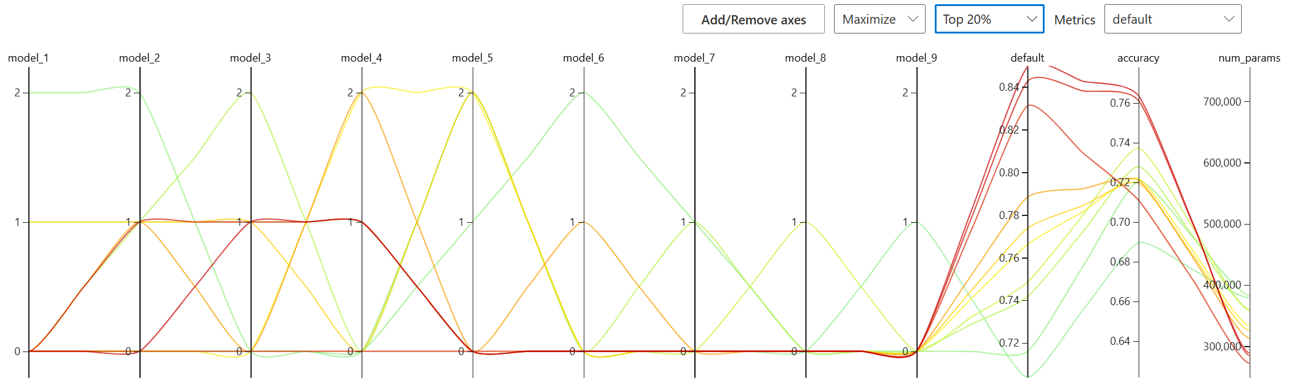}
    \caption{Performance of top 20\% trials of Policy-based RL.}
    \label{fig:Policy-basedRL_top20}
\end{figure}

\begin{figure}[h]
    \centering
    \includegraphics[width=1\textwidth]{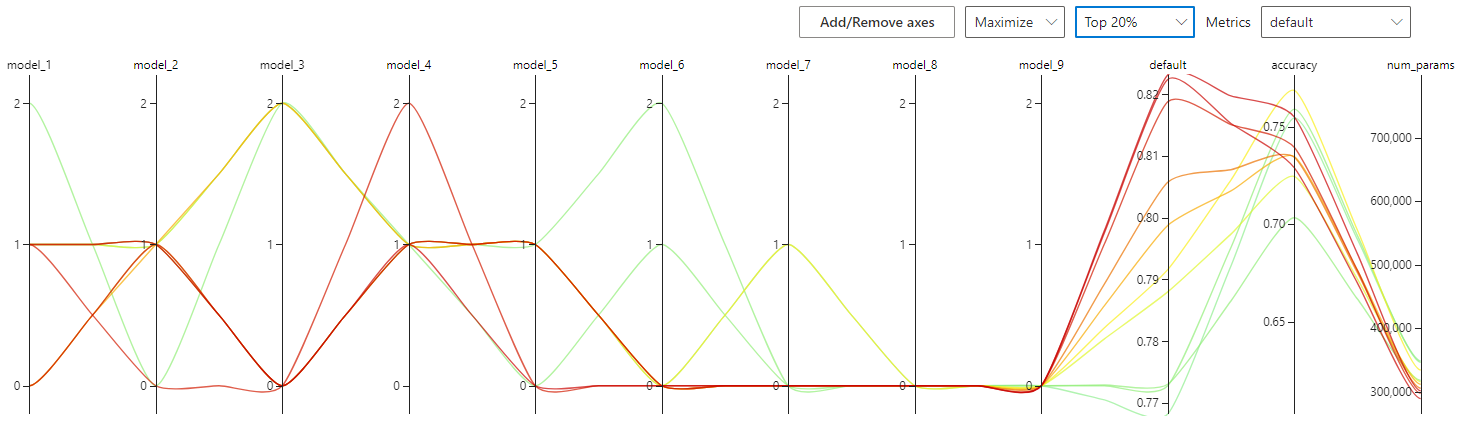}
    \caption{Performance of top 20\% trials of Regularized Evolution.}
    \label{fig:regularized_top20}
\end{figure}

\begin{figure}[h]
    \centering
    \includegraphics[width=1\textwidth]{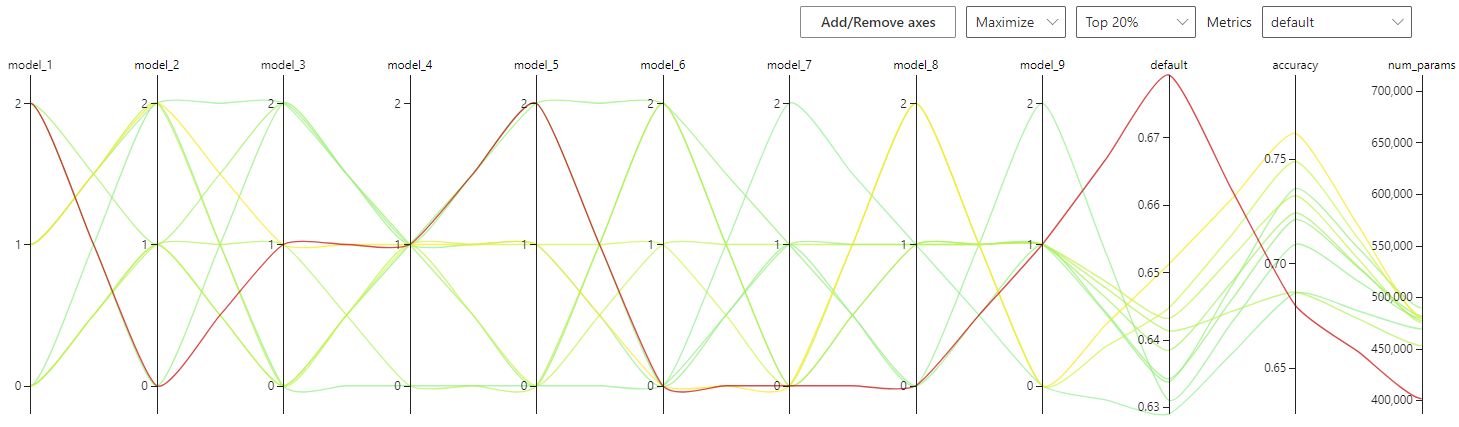}
    \caption{Performance of top 20\% trials of TPE.}
    \label{fig:tpe_top20}
\end{figure}

\begin{figure}[h]
    \centering
    \includegraphics[width=1\textwidth]{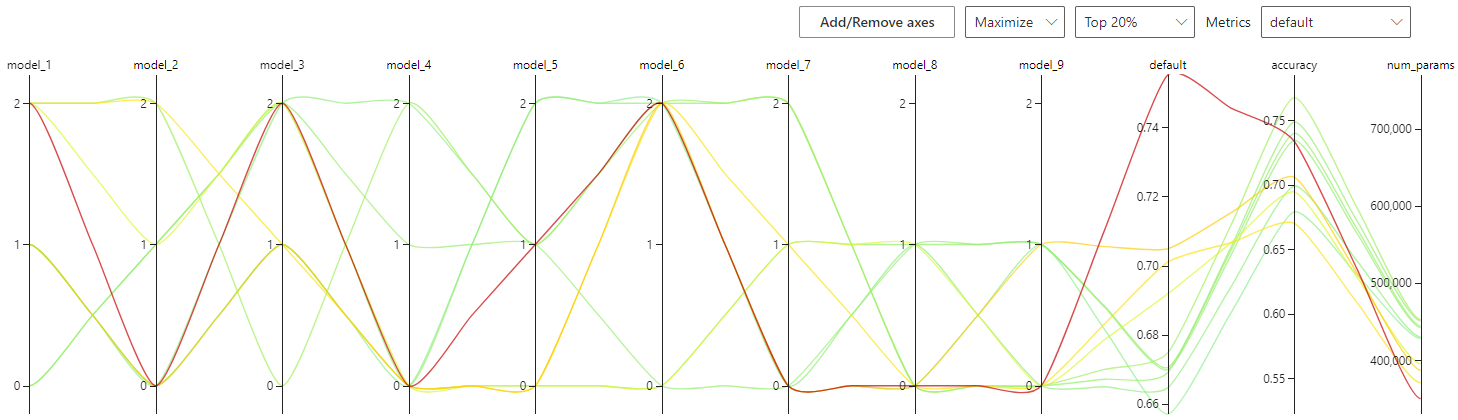}
    \caption{Performance of top 20\% trials of Multi-trial Random.}
    \label{fig:random_top20}
\end{figure}

Our analysis of the top 20\% of models revealed interesting trade-offs between accuracy and model size. The performance of top 20\% trials for Policy-based RL, Regularized Evolution, TPE, and Multi-trial Random are shown respectively in Figures \ref{fig:Policy-basedRL_top20}, \ref{fig:regularized_top20}, \ref{fig:tpe_top20}, and \ref{fig:random_top20}.

\begin{enumerate}
    \item Policy-based RL consistently produced models with a good balance between accuracy and size, as reflected in its high \textit{M-Factor} values. This suggests that the RL agent learned to optimize for both aspects effectively. The consistency in performance indicates that the learned policy was robust and reliably produced well-balanced architectures.

    \item Regularized Evolution showed a notable trend of reducing model size after the 30th trial while maintaining competitive accuracy. This behavior demonstrates the strategy's ability to refine solutions over time, progressively finding architectures that maintain high accuracy with increased efficiency. It suggests that Regularized Evolution may be particularly effective for tasks where model efficiency is a critical concern.

    \item TPE and Multi-trial Random showed more variability in both accuracy and model size. This variability reflects their less directed search processes. For TPE, it suggests that the algorithm struggled to build a consistent model of the relationship between architectural choices and our custom metric. For Multi-trial Random, the variability is expected due to the nature of random sampling.
\end{enumerate}

Policy-based RL and Regularized Evolution stand out for their high consistency and good to excellent balance between model size and accuracy. The fast optimization speed of Regularized Evolution is particularly noteworthy, as it suggests this strategy could be valuable in resource-constrained scenarios.

In conclusion, our results demonstrate the effectiveness of Policy-based RL and Regularized Evolution for NAS in the context of ResNet architectures on CIFAR-10, when optimizing for both accuracy and model size. The study underscores the importance of choosing appropriate search strategies and evaluation metrics in NAS. The strong performance of these methods, coupled with random search, provides valuable insights for researchers and practitioners in the field of neural architecture search.

\section{Conclusion and Future Work}
This paper introduces the \textit{M-factor} metric to aid in the process of searching for the desired ResNet architecture in resource-constrained environments using NAS. We also compare the performance of four different NAS techniques. Policy-based Reinforcement Learning achieved an \textit{M-factor} value of 0.84, with performance improvements observed after the 39th trial. Regularized Evolution reached an \textit{M-factor} of 0.82, showing initial optimization within 20 trials and subsequent model size reduction. Multi-trial Random search attained an \textit{M-factor} of 0.75, surpassing the Tree-structured Parzen Estimator's 0.67 in the defined ResNet-based search space for CIFAR-10. This search space comprised three layers, each with three ResNet blocks, allowing for 19,683 possible architectures through variations in convolutional layer configurations. These results indicate variations in strategy performance based on computational budget and optimization goals. The study demonstrates the potential of tailored metrics in guiding NAS towards architectures balancing accuracy and efficiency.

Future research directions include examining the \textit{M-factor}'s application to expanded search spaces and model architectures, and exploring its integration with one-shot NAS methods. Extending the study to different datasets and tasks will assess the generalizability of the \textit{M-factor} metric and the performance of various NAS strategies across domains. Incorporating additional efficiency metrics, such as inference time or energy consumption, into the \textit{M-factor} may provide insights for specific deployment scenarios.


\bibliographystyle{splncs04}
\bibliography{references.bib}

\begin{thebibliography}{10}
\providecommand{\url}[1]{\texttt{#1}}
\providecommand{\urlprefix}{URL }
\providecommand{\doi}[1]{https://doi.org/#1}

\bibitem{baker2016designing}
Baker, B., Gupta, O., Naik, N., Raskar, R.: Designing neural network architectures using reinforcement learning. arXiv preprint arXiv:1611.02167  (2016)

\bibitem{baker2022designing}
Baker, B., Gupta, O., Naik, N., Raskar, R.: Designing neural network architectures using reinforcement learning. In: International Conference on Learning Representations (2022)

\bibitem{bender2018understanding}
Bender, G., Kindermans, P.J., Zoph, B., Vasudevan, V., Le, Q.: Understanding and simplifying one-shot architecture search. In: International Conference on Machine Learning. pp. 550--559 (2018)

\bibitem{bergstra2012random}
Bergstra, J., Bengio, Y.: Random search for hyper-parameter optimization. Journal of machine learning research  \textbf{13}(2),  281--305 (2012)

\bibitem{bergstra2013making}
Bergstra, J., Yamins, D., Cox, D.D.: Making a science of model search: Hyperparameter optimization in hundreds of dimensions for vision architectures. International conference on machine learning pp. 115--123 (2013)

\bibitem{bergstra2011algorithms}
Bergstra, J.S., Bardenet, R., Bengio, Y., K{\'e}gl, B.: Algorithms for hyper-parameter optimization. In: Advances in neural information processing systems. pp. 2546--2554 (2011)

\bibitem{cai2019proxylessnas}
Cai, H., Zhu, L., Han, S.: Proxylessnas: Direct neural architecture search on target task and hardware. In: International Conference on Learning Representations (2019)

\bibitem{canziani2016analysis}
Canziani, A., Paszke, A., Culurciello, E.: An analysis of deep neural network models for practical applications. arXiv preprint arXiv:1605.07678  (2016)

\bibitem{chitty2023neural}
Chitty-Venkata, K.T., Emani, M., Vishwanath, V., Somani, A.K.: Neural architecture search benchmarks: Insights and survey. IEEE Access  \textbf{11},  25217--25236 (2023)

\bibitem{elsken2019neural}
Elsken, T., Metzen, J.H., Hutter, F.: Neural architecture search: A survey. Journal of Machine Learning Research  \textbf{20}(55),  1--21 (2019)

\bibitem{fang2019tinier}
Fang, W., Wang, L., Ren, P.: Tinier-yolo: A real-time object detection method for constrained environments. Ieee Access  \textbf{8},  1935--1944 (2019)

\bibitem{ferger1931nature}
Ferger, W.F.: The nature and use of the harmonic mean. Journal of the American Statistical Association  \textbf{26}(173),  36--40 (1931)

\bibitem{he2016deep}
He, K., Zhang, X., Ren, S., Sun, J.: Deep residual learning for image recognition. In: Proceedings of the IEEE conference on computer vision and pattern recognition. pp. 770--778 (2016)

\bibitem{10.1145/3665138}
Heuillet, A., Nasser, A., Arioui, H., Tabia, H.: Efficient automation of neural network design: A survey on differentiable neural architecture search. ACM Comput. Surv.  \textbf{56}(11) (jun 2024). \doi{10.1145/3665138}, \url{https://doi.org/10.1145/3665138}

\bibitem{howard2017mobilenets}
Howard, A.G., Zhu, M., Chen, B., Kalenichenko, D., Wang, W., Weyand, T., Andreetto, M., Adam, H.: Mobilenets: Efficient convolutional neural networks for mobile vision applications. In: Proceedings of the IEEE conference on computer vision and pattern recognition. pp. 4510--4520 (2017)

\bibitem{kang2023neural}
Kang, J.S., Kang, J., Kim, J.J., Jeon, K.W., Chung, H.J., Park, B.H.: Neural architecture search survey: A computer vision perspective. Sensors  \textbf{23}(3), ~1713 (2023)

\bibitem{krizhevsky2009learning}
Krizhevsky, A., Hinton, G., et~al.: Learning multiple layers of features from tiny images  (2009)

\bibitem{li2020random}
Li, L., Talwalkar, A.: Random search and reproducibility for neural architecture search. In: Uncertainty in Artificial Intelligence. pp. 367--377. PMLR (2020)

\bibitem{liu2018progressive}
Liu, C., Zoph, B., Neumann, M., Shlens, J., Hua, W., Li, L.J., Fei-Fei, L., Yuille, A., Huang, J., Murphy, K.: Progressive neural architecture search. In: Proceedings of the European Conference on Computer Vision (ECCV). pp. 19--34 (2018)

\bibitem{liu2017hierarchical}
Liu, H., Simonyan, K., Vinyals, O., Fernando, C., Kavukcuoglu, K.: Hierarchical representations for efficient architecture search. arXiv preprint arXiv:1711.00436  (2017)

\bibitem{liu2018darts}
Liu, H., Simonyan, K., Yang, Y.: Darts: Differentiable architecture search. In: International Conference on Learning Representations (2019)

\bibitem{lopes2024manas}
Lopes, V., Carlucci, F.M., Esperan{\c{c}}a, P.M., Singh, M., Yang, A., Gabillon, V., Xu, H., Chen, Z., Wang, J.: Manas: multi-agent neural architecture search. Machine Learning  \textbf{113}(1),  73--96 (2024)

\bibitem{lyu2024survey}
Lyu, Z., Yu, T., Pan, F., Zhang, Y., Luo, J., Zhang, D., Chen, Y., Zhang, B., Li, G.: A survey of model compression strategies for object detection. Multimedia Tools and Applications  \textbf{83}(16),  48165--48236 (2024)

\bibitem{mellor2021neural}
Mellor, J., Turner, J., Storkey, A., Crowley, E.J.: Neural architecture search without training. In: International conference on machine learning. pp. 7588--7598. PMLR (2021)

\bibitem{miikkulainen2024evolving}
Miikkulainen, R., Liang, J., Meyerson, E., Rawal, A., Fink, D., Francon, O., Raju, B., Shahrzad, H., Navruzyan, A., Duffy, N., et~al.: Evolving deep neural networks. In: Artificial intelligence in the age of neural networks and brain computing, pp. 269--287. Elsevier (2024)

\bibitem{paszke2019pytorch}
Paszke, A., Gross, S., Massa, F., Lerer, A., Bradbury, J., Chanan, G., Killeen, T., Lin, Z., Gimelshein, N., Antiga, L., et~al.: Pytorch: An imperative style, high-performance deep learning library. In: Advances in neural information processing systems. pp. 8026--8037 (2019)

\bibitem{pham2018efficient}
Pham, H., Guan, M.Y., Zoph, B., Le, Q.V., Dean, J.: Efficient neural architecture search via parameter sharing. In: International Conference on Machine Learning. pp. 4095--4104 (2018)

\bibitem{real2019regularized}
Real, E., Aggarwal, A., Huang, Y., Le, Q.V.: Regularized evolution for image classifier architecture search. In: Proceedings of the aaai conference on artificial intelligence. vol.~33, pp. 4780--4789 (2019)

\bibitem{real2020automl}
Real, E., Liang, C., So, D.R., Le, Q.V.: Automl-zero: Evolving machine learning algorithms from scratch. arXiv preprint arXiv:2003.03384  (2020)

\bibitem{ren2021comprehensive}
Ren, P., Xiao, Y., Chang, X., Huang, P.Y., Li, Z., Chen, X., Wang, X.: A comprehensive survey of neural architecture search: Challenges and solutions. ACM Computing Surveys (CSUR)  \textbf{54}(4),  1--34 (2021)

\bibitem{tan2019efficientnet}
Tan, M., Le, Q.V.: Efficientnet: Rethinking model scaling for convolutional neural networks. arXiv preprint arXiv:1905.11946  (2019)

\bibitem{wong2019netscore}
Wong, A.: Netscore: towards universal metrics for large-scale performance analysis of deep neural networks for practical on-device edge usage. In: International Conference on Image Analysis and Recognition. pp. 15--26. Springer (2019)

\bibitem{wu2019fbnet}
Wu, B., Dai, X., Zhang, P., Wang, Y., Sun, F., Wu, Y., Tian, Y., Vajda, P., Jia, Y., Keutzer, K.: Fbnet: Hardware-aware efficient convnet design via differentiable neural architecture search. In: Proceedings of the IEEE/CVF Conference on Computer Vision and Pattern Recognition. pp. 10734--10742 (2019)

\bibitem{zhang2020memory}
Zhang, H., Li, Y., Chen, H., Shen, C.: Memory-efficient hierarchical neural architecture search for image denoising. In: Proceedings of the IEEE/CVF conference on computer vision and pattern recognition. pp. 3657--3666 (2020)

\bibitem{zoph2016neural}
Zoph, B., Le, Q.V.: Neural architecture search with reinforcement learning. arXiv preprint arXiv:1611.01578  (2016)

\bibitem{zoph2017neural}
Zoph, B., Le, Q.V.: Neural architecture search with reinforcement learning. In: International Conference on Learning Representations (2017)

\bibitem{zoph2018learning}
Zoph, B., Vasudevan, V., Shlens, J., Le, Q.V.: Learning transferable architectures for scalable image recognition. In: Proceedings of the IEEE conference on computer vision and pattern recognition. pp. 8697--8710 (2018)

\end{thebibliography}
\end{document}